\pdfoutput=1

\documentclass[11pt]{article}

\usepackage{ACL2023}

\usepackage{times}
\usepackage{latexsym}

\usepackage[T1]{fontenc}

\usepackage[utf8]{inputenc}

\usepackage{microtype}

\usepackage{amssymb}
\usepackage{amsmath}
\usepackage{tabularx}
\usepackage{multirow}
\usepackage{caption}
\usepackage{subcaption}
\usepackage{colorprofiles}
\usepackage[capitalise]{cleveref}
\usepackage{adjustbox}
\usepackage{float}
\usepackage{xspace}
\usepackage{booktabs}
\usepackage{nicefrac}
\usepackage{graphicx}
\graphicspath{ {./graphs/} }
\usepackage{pgf}
\usepackage{dblfloatfix}
\usepackage[colorinlistoftodos]{todonotes}

\usepackage{inconsolata}

\newcommand{\com}[1]{}
\newcommand{\resolved}[1]{}

\newcommand{\draftcomment}[3]{{\textcolor{#3}{[#1]#2}}}

\newcommand{\roy}[1]{\draftcomment{#1}{\textsc{Roy}}{orange}}
\newcommand{\rr}[2]{\roy{\sout{#1} #2}}

\newcommand{\rs}[2]{\roy{\sout{#1}}}

\renewcommand{\rs}[1]{}
\renewcommand{\rr}[2]{#2}

\newcommand{\methodname}{SWEET\xspace}
\newcommand{\MM}{Multi-Model\xspace}
\newcommand{\EE}{Early-Exit\xspace}
\newcommand{\bert}{\(\text{BERT}\)\xspace}
\newcommand{\bertbase}{\(\text{BERT}_{\text{\textit{BASE}}}\)\xspace}
\newcommand{\bertlarge}{\(\text{BERT}_{\text{\textit{LARGE}}}\)\xspace}

\newcommand{\deberta}{\(\text{DeBERTa}\)\xspace}

\title{Finding the SWEET Spot:\\ Analysis and Improvement of Adaptive Inference in Low Resource Settings}

\author{\textbf{Daniel Rotem}$^\heartsuit$ \quad\textbf{Michael Hassid}$^{\heartsuit}$ \quad
\textbf{Jonathan Mamou}$^\spadesuit$ \quad
\textbf{Roy Schwartz}$^\heartsuit$ \\
  $^\heartsuit$School of Computer Science \& Engineering, Hebrew University of Jerusalem\\
  $^\spadesuit$ Intel Labs, Israel\\
{\tt \{daniel.rotem,michael.hassid,roy.schwartz1\}@mail.huji.ac.il}\\
  {\tt jonathan.mamou@intel.com}
  }

\begin{document}
\maketitle
\begin{abstract}
Adaptive inference is a simple method for reducing inference costs. The method works by maintaining multiple classifiers of different capacities, and allocating resources to each test instance according to its difficulty. In this work, we compare the two main approaches for adaptive inference, \EE and \MM, when training data is limited. First, we observe that for models with the same architecture and size, individual \MM classifiers outperform their \EE counterparts by an average of \(2.3\%\). We \rr{provide empirical evidence}{show} that this gap is caused by \EE classifiers sharing model parameters during training, resulting in conflicting gradient updates of model weights. We find that despite this gap, \EE still provides a better speed-accuracy trade-off due to the overhead of the \MM approach. To address these issues, we propose \methodname,\footnote{\textbf{S}eparating \textbf{W}eights in \textbf{E}arly \textbf{E}xit \textbf{T}ransformers.} an \EE fine-tuning method that assigns each classifier its own set of unique model weights, not updated by other classifiers. We compare \methodname's speed-accuracy curve to standard \EE and \MM baselines and find that it outperforms both methods at fast speeds while maintaining comparable \rr{performance}{scores} to \EE at slow speeds. Moreover, \methodname individual classifiers outperform \EE ones by \(1.1\%\) on average\rs{mainly improving performance of earlier classifiers}. \methodname enjoys the benefits of both methods, paving the way for further reduction of inference costs in NLP. We publicly release our code.\footnote{\url{https://github.com/schwartz-lab-NLP/SWEET}}
 
\end{abstract}
\section{Introduction}
\label{sec:introduction}
\begin{figure}[t]
\centering
\includegraphics[trim={0 0 0 0}, clip, width=\columnwidth]{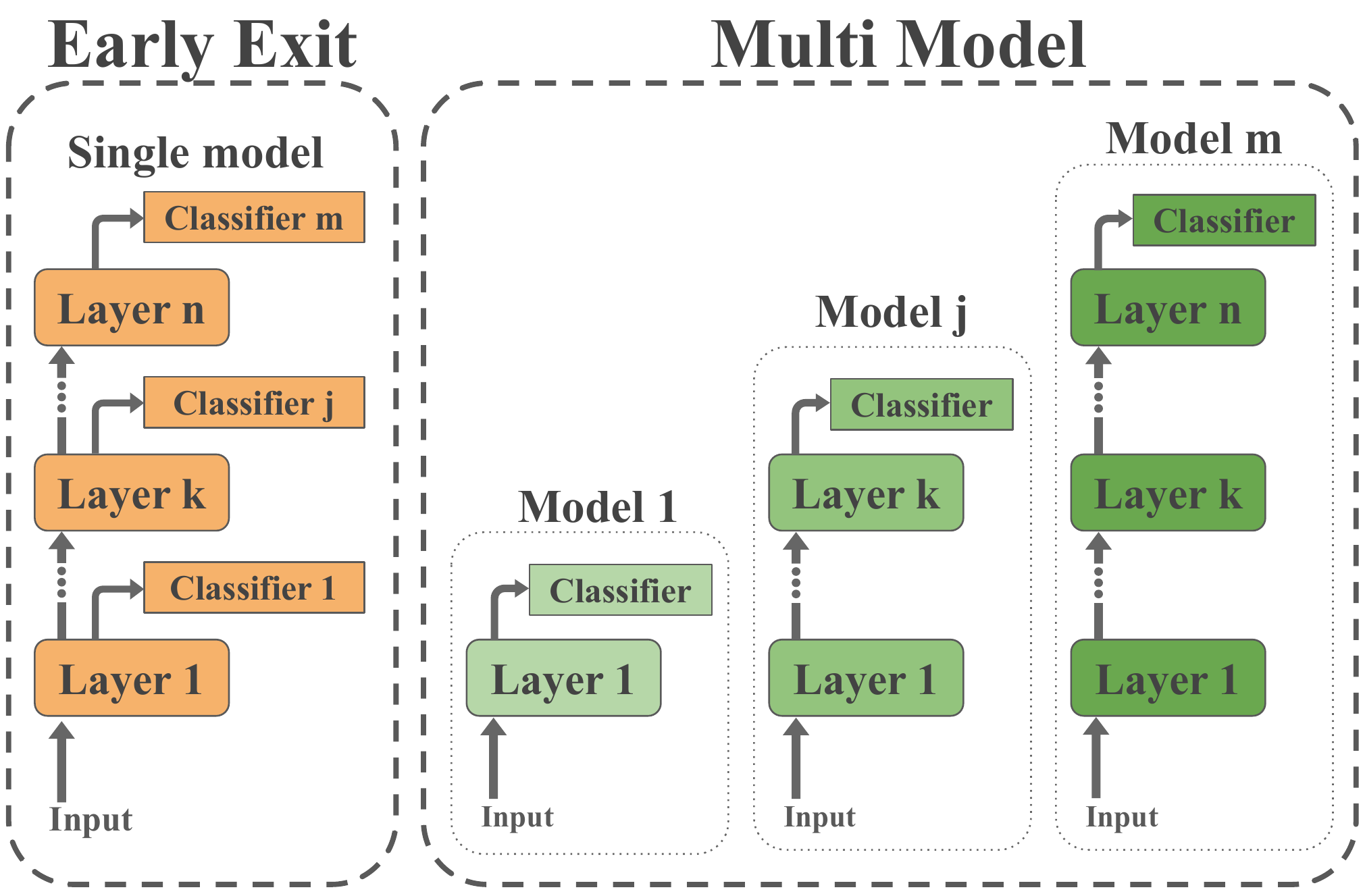}

\caption{\label{schwartz figure}
Illustration of the adaptive inference approaches compared in this work. In both methods, multiple classifiers of increasing sizes are run serially, until a confident prediction is made. In \EE (left), a single model with multiple classifiers is used, such that early computations are reused by later classifiers. In \MM (right), a sequence of independent models is used, allowing each classifier to decouple its parameters from other classifiers.}
\end{figure}

Pre-trained Transformer-based {language models} such as \bert~\cite{devlin-etal-2019-bert}, DeBERTa~\cite{he2020deberta}, and GPT3~\cite{brown2020language} have become the go-to tool in NLP . Although powerful, the growing size of these models has been a major drawback~\cite{thompson2020computational, schwartz2020green}, making them costly to run. Various attempts to reduce inference cost have been proposed, including distillation~\cite{hinton2015distillation}, pruning~\cite{lecun1989optimal} and quantization~\cite{courbariaux2014training}. This work focuses on adaptive inference~\cite{graves2016adaptive, liu-etal-2020-fastbert}, a recent approach in which the variability of sample difficulty is leveraged toward a smarter allocation of computational resources. An appealing property of adaptive inference is that it enables dynamic control of the speed-accuracy trade-off.

There are two main approaches to adaptive inference, both using a set of classifiers of different sizes. In \textit{Early-Exit} 
\cite{schwartz-etal-2020-right,xin-etal-2020-deebert}, multiple classification heads are added to the same model at different layers, allowing for early exit during inference (\cref{schwartz figure}, left). Another approach (henceforth \textit{Multi-Model}) is to apply multiple  \textit{independent} classifiers of varying capacities serially until a prediction is made (\citealp{varshney2022model, li2020cascadebert}, \cref{schwartz figure}, right).
These approaches have complementary benefits; \MM allows for easier batching at inference time, and potentially larger savings due to using very efficient models~\cite{mamou2022tangobert}. \EE on the other hand is more memory efficient, faster to train, and enables re-use of early computation if an early exit was not taken.

\begin{figure}[t]
\centering
\includegraphics[trim={0 0 0 0}, clip, width=\columnwidth]{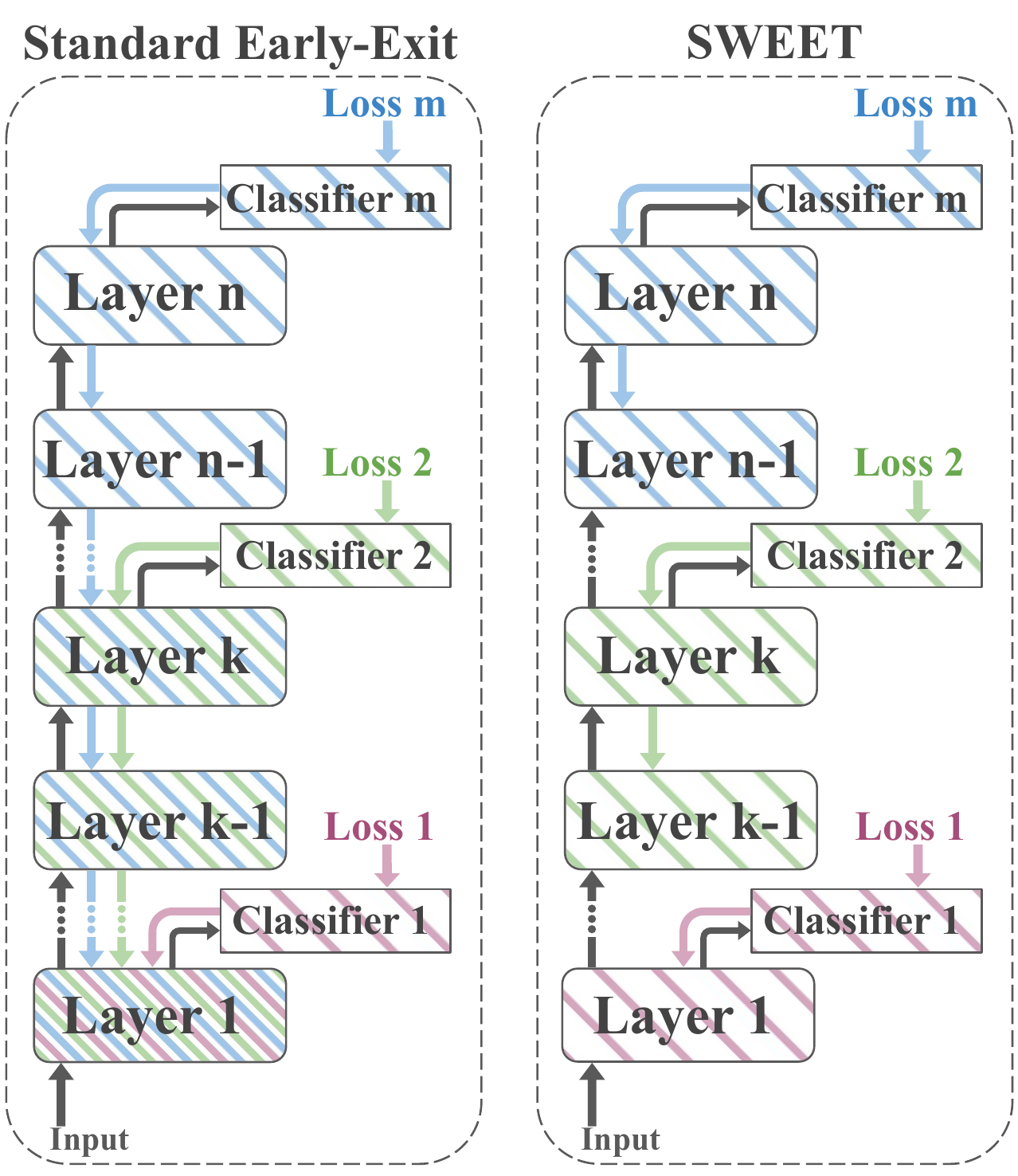}
\caption{\label{SWEET}
\textbf{Left}: standard \EE fine-tuning, where lower layers get gradient updates from multiple classifiers. \textbf{Right}: our \methodname method, in which each layer parameters are updated only by the next classifier.}
\end{figure}

In this work, we compare the speed-accuracy behavior of the two approaches when training data is limited.\footnote{Reducing inference costs is particularly helpful when computational resources are limited. Such conditions are often paired with restricted access to labeled data or a low budget for data annotation. To evaluate the effectiveness of adaptive inference methods in such scenarios, we limit training data size to a few thousand examples in all our experiments.}
We first observe that \EE model weights are updated by multiple conflicting gradient signals throughout the training process (\cref{SWEET}, left). We show that this leads to a decrease in performance of \textit{individual} \EE classifiers compared to \MM ones (\(2.3\%\) gap on average). We find that this gap is higher for the earliest classifiers (\(5.2\%\) on average) than for later ones (\(1.4\%\)). 

We also find that while each \MM classifier outperforms its \EE counterpart, it does not translate to an overall better speed-accuracy trade-off. Instead, we find that each method dominates performance in a different region: \MM  outperforms \EE  at fast inference speeds,  while \EE is better at slow speeds. \MM downgraded scores at slow speeds are likely caused by the overhead of running models sequentially to predict hard samples.

Inspired by our findings, we present \methodname,\footnote{\textbf{S}eparating \textbf{W}eights for \textbf{E}arly-\textbf{E}xit \textbf{T}ransformers.} an \EE method for bridging the performance gap between standard \EE and \MM. In \methodname, each \EE classifier only updates the parameters of layers preceding it up to the previous classifier. This way, each model parameter is updated by a single classifier, thus avoiding conflicting gradients during the training of \EE models (\cref{SWEET}, right). 

We experiment with two established pre-trained models: \bert~\cite{devlin-etal-2019-bert} and \deberta~\cite{he2020deberta}. We fine-tune \EE models using \methodname on seven text classification tasks from GLUE~\cite{wang-etal-2018-glue} and compare them to \EE and \MM baselines. The speed-accuracy curve of \methodname dominates both baselines at fast speeds for \(21\) out of \(28\) experiments conducted. As for individual classifiers, \methodname performs \(1.1\%\) better on average than \EE, mostly improving earlier classifiers, where conflicting gradients are more dominant.

We summarise our main contributions: \textbf{(1)} We propose a way of measuring conflicting gradients, and show that they exist in \EE training process; \textbf{(2)} We empirically compare \EE and \MM classifiers and show that conflicting gradients lead to individual \EE classifiers being less accurate; \textbf{(3)} We propose a novel fine-tuning method, \methodname, which alleviates the conflicting gradients problem in \EE, and leads to improved results at fast inference speeds; \textbf{(4)} We publicly release our code.\footnote{\url{{https://github.com/schwartz-lab-NLP/SWEET}}}

\section{Background: Adaptive Inference}
\label{sec:Background}
Adaptive inference aims to reduce inference costs of deep neural nets by matching model and sample complexities. Sample difficulty usually varies in real-world data, meaning not all instances require processing by the most powerful classifier. Therefore, we can allocate fewer resources to easier instances, and reduce the average inference cost, potentially at the cost of performance degradation. 

Several exit strategies have been developed to control the speed-accuracy trade-off of a model by deciding when to make an early prediction and halt computation \cite{xin-etal-2020-deebert,  zhou2020bert,  xin-etal-2021-berxit, schuster2021consistent, zhang2022pcee}. In this work, we mainly experiment with confidence-based exiting \cite{schwartz-etal-2020-right}, in which computation halts if the softmax probability assigned to a given label exceeds a predetermined threshold. Dynamic control of model's inference speed is gained through setting different threshold values. There are two main adaptive inference approaches, \EE and \MM, which we describe below.

\paragraph{\EE}
\EE models are deep neural nets with multiple output points, following intermediate layers. In this work, we focus on \EE implementation as presented in \citet{schwartz-etal-2020-right}. During fine-tuning,\footnote{Some works have pre-trained \EE Models from scratch \cite{liu2021towards} but due to budgetary constraints, it is common practice to fine-tune pre-trained models with the added classifiers on the downstream task \cite{liu-etal-2020-fastbert,xin-etal-2020-deebert,schwartz-etal-2020-right}.} instances are passed through the model, and a loss is calculated based on predictions made by all classifiers. This leads to some model weights being updated by gradient signals from multiple classifiers (\cref{SWEET}, left).

At inference time, an instance is passed through the model and its label is predicted sequentially until a decision to exit early is taken, and computation is halted. An appealing property of \EE is that it allows for efficient re-use of previous computation from lower layers in higher classifiers. However, this means that some model parameters are updated by multiple classifiers, which may lead to sub-optimal performance of individual classifiers due to conflicting gradients. In this work, we study the effect of this property on \EE models.

\paragraph{\MM}
In \MM adaptive inference, a set of independent models of increasing capacity are fine-tuned separately for the same task. At inference time, the models are used sequentially from smallest to largest until a prediction meets some predetermined criterion or until the largest model has been used. This method is more robust than \EE, being easier to extend and enabling the use of different architectures. Additionally, \MM potentially allows for further computational savings by using models smaller than the smallest \EE model (a single layer of the backbone model, \citealp{mamou2022tangobert}). However, it may add overhead to the prediction time of hard instances, as those pass through multiple models with early computations being discarded, bringing the total runtime to exceed that of using the largest model.

\section{\EE vs.~\MM}
\label{sec:Motivation}

\subsection{Conflicting Gradients in \EE \label{conflicting_gradients}}
Unlike \MM, when fine-tuning \EE models, model weights are updated by multiple gradient signals, originating in different classification layers (\cref{SWEET}, left). We hypothesize that this leads to sub-optimal performance for all classifiers involved, as gradient signals might conflict with one another and derail the classifiers from their goal. To test this hypothesis, we compare the gradient similarity of different \EE classifiers.

We fine-tune a \bertbase model with four exit points (following layers \(\left[1,4,6,12\right]\)) for \(400\) training steps on the MNLI dataset~\cite{williams-etal-2018-broad}, using a batch size of 16 with a learning rate of 2e-5. We pass a new training batch through the model,\footnote{We repeat this experiment with an additional batch. Results (\cref{sec:appendix_B}) show a very similar trend.} and inspect the gradients with respect to the last feed-forward matrix in each layer preceding a classifier.\footnote{Except for the last layer, updated by a single classifier.}
To measure the degree of alignment between gradient updates of different classifiers, we average the cosine similarity between the rows of gradient matrices for every pair of classifiers. High similarity indicates that the classifiers are updating the weights in a similar direction, while low similarity (in absolute values)  suggests that the updates are close to orthogonal and potentially detrimental to both classifiers.

\begin{figure}[t]
\centering
\includegraphics[trim={0 0 0 0}, clip, width=\columnwidth]{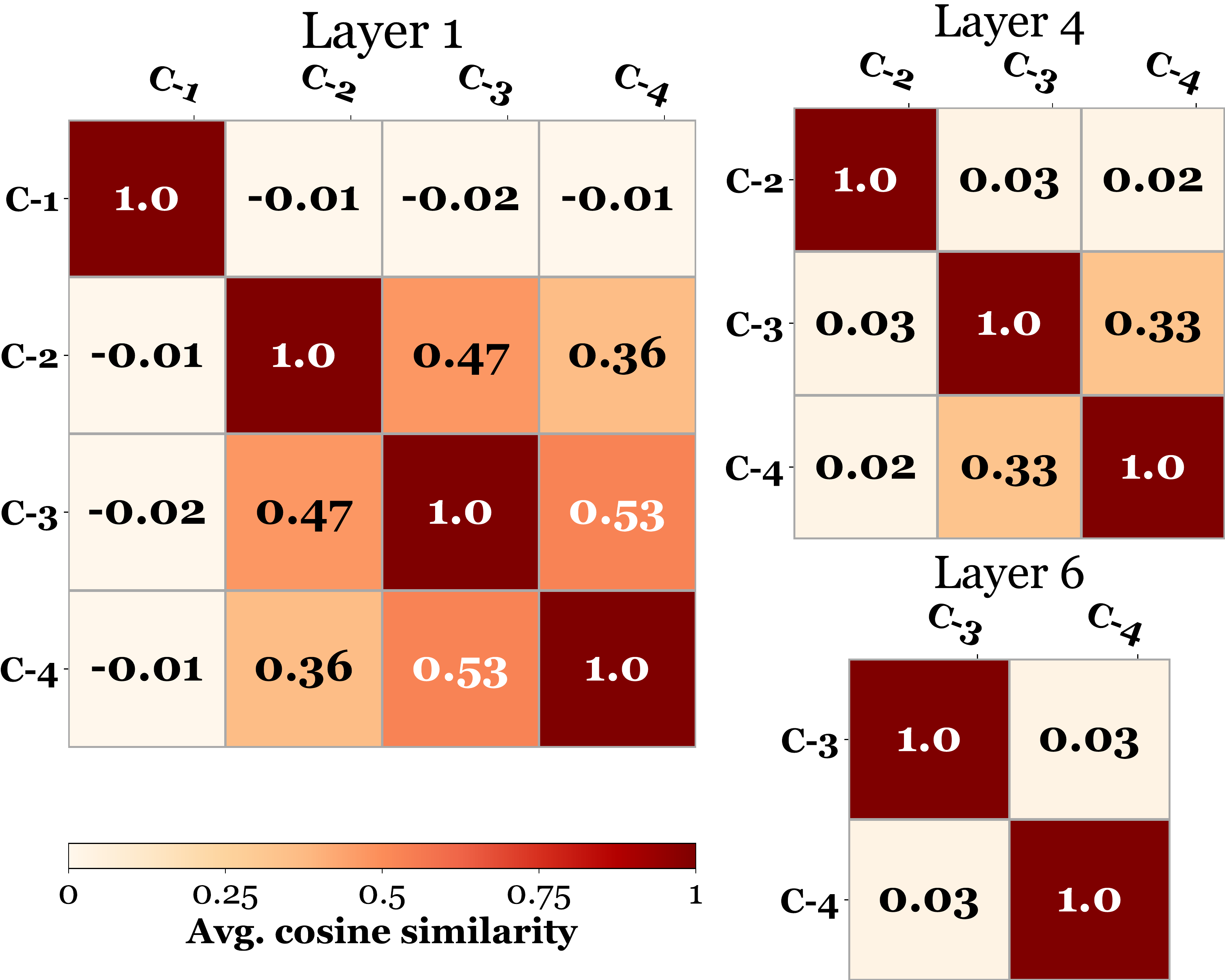}
\caption{\label{gradient_similarity}
Average cosine similarity between classifiers' gradient updates of model layers. C-i stands for Classifier \(i\). Layer 1 (preceding C-1) is updated by 3 classifiers, while layers 4 (preceding  C-2), and 6 (C-3) are updated by 3 \& 2 classifiers respectively.
For each layer, the gradient update of the following classifier is roughly orthogonal to those of future classifiers, whereas gradient updates of higher classifiers tend to better align with one another.}
\end{figure}

This section studies the strengths and weaknesses of both adaptive inference approaches. We start by describing a limitation of \EE approaches: lower model layers are updated by conflicting gradient signals during training. We show that this leads to inferior performance of individual \EE classifiers compared to corresponding \MM classifiers. We then compare the effects of these performance drops on the speed-accuracy curve of \EE models compared \MM ones.

\cref{gradient_similarity} shows
that the gradients of future classifiers are generally orthogonal to those of the current classifier for each examined Transformer block. This indicates that model weights indeed suffer from conflicting gradient updates during the training process, which might affect the performance of each individual classifier. Interestingly, when multiple future classifiers are present, they tend to align with each other, as indicated by the relatively high similarity between layer 1's gradient updates originating in classifiers 2, 3, and 4, and between layer 4's gradient updates from classifiers 3 and 4.

\subsection{Effects of Conflicting Gradients}{\label{paragraph: effects CG}}

To evaluate the effect of conflicting gradients on individual \EE classifiers, we compare the different classifiers of an \EE model and a \MM model. For a clean comparison, we use the same backbone model, exit points, hyper-parameters,\footnote{Except for the learning rate, which was tuned for each task and model individually.} and random seeds. As an example, for a given 12-layer Transformer model, an \EE model with exit points following layers \([4,12]\) would be compared to a \MM one consisting of two classifiers: the first four layers fit with a classification head, and a full (12 layer) model. The models differ only in the fact that for the \EE model, model weights are updated by multiple future classifiers during the fine-tuning process, while each \MM classifier is an independent model. To isolate the effect of conflicting gradients on the classifiers, we evaluate each one separately on the entire validation set. The performance gap between each individual \MM classifier and its corresponding \EE classifier, allows us to directly measure the latter's downgrade in performance caused by conflicting gradients.

\setlength{\tabcolsep}{2.3pt} 

\begin{table}[t] 
\centering

\begin{tabularx}{\columnwidth}{@{} lccccc@{}}
\toprule
 \textbf{Size}&\textbf{Method} & \multicolumn{4}{c}{\textbf{Exit Layer}}\\[0.1ex]
 \midrule\rule{0pt}{1.0ex}
&&\textbf{1} & \textbf{4} & \textbf{6} & \textbf{12}\\[0.1ex]
\cmidrule{3-6}
\multirow{3}{*}{BASE} 
& MM &  \(\textbf{60.9}_{0.6}\) & \(71.4_{0.1}\) & \(\textbf{74.7}_{1.2}\) & \(\textbf{79.9}_{0.9}\) \\ & EE & \(57.3_{0.3}\) & \(70.3 _{0.3}\) & \(74.4_{0.7}\) & \(78.7_{0.5}\) \\
& \methodname & \(\textbf{60.9}_{0.5}\) & \(\textbf{71.6} _{0.5}\) & \(74.0_{0.4}\) & \(77.4_{0.6}\)\\ [1ex]
\midrule
&&\textbf{1} & \textbf{6} & \textbf{12} & \textbf{24}\\
\cmidrule{3-6}
\multirow{3}{*}{LARGE}& MM &  \(\textbf{60.1}_{0.1}\) & \(\textbf{66.9}_{0.4}\) & \(74.4_{0.9}\) & \(\textbf{81.6}_{0.3}\)\\
& EE & \(56.6_{0.3}\) & \(65.6_{0.9}\) & \(74.0_{0.6}\) & \(79.9_{1.9}\) \\
& \methodname & \(59.8_{0.3}\) & \(66.5_{0.7}\) & \(\textbf{74.5}_{0.3}\) & \(81.3_{1.0}\) \\
\bottomrule
\end{tabularx} 
\caption{\label{tab:layer_comparison} 
Results of individual classification layers averaged across all tasks using \bert as a backbone model. \MM (\textbf{MM}) classifiers outperform their \EE (\textbf{EE}) counterparts, with the gap being the largest for early classifiers. \methodname closes much of this gap, especially for early classifiers.  Standard deviation (across random seeds) is reported in subscript. Results for \deberta are presented in \cref{tab:deberta_layer_comparison}.}

\end{table}
We experiment with \bert and \deberta \{BASE (\({\sim}110\)M parameters), and LARGE (\({\sim}350\)M parameters)\}. For BASE versions, we install classifiers following layers \(\left[1,4,6,12\right]\). For LARGE versions, we use layers \(\left[1,6,12,24\right]\). We fine-tune an \EE model and a corresponding \MM model on seven NLP tasks from the GLUE benchmark~\cite{GLUE}: SST-2~\cite{socher-etal-2013-recursive}, MRPC~\cite{dolan-brockett-2005-automatically}, RTE~\cite{dagan2006pascal, haim2006second, giampiccolo2007third, bentivogli2009fifth}, CoLA~\cite{warstadt-etal-2019-neural}, MNLI~\cite{williams-etal-2018-broad}, QNLI~\cite{rajpurkar2016squad}, and QQP~\cite{qqp}. We report accuracy for all tasks except for CoLA (Matthews corr.).
We fine-tune the models for two epochs on all tasks. As our goal is to test Adaptive inference in a low resource setting, we limit the training set size for each task to \(6K\).\footnote{For datasets smaller than $6K$, we use the entire dataset. The training sets are randomly sampled for each random seed, and are shared across methods with the same seed.}  We report the mean validation scores and standard deviation across three random seeds.
Other hyper-parameters are listed in \cref{sec:appendix_A}.

\cref{tab:layer_comparison} shows results for \bert classifiers,\footnote{
 See \cref{sec:appendix_C}, \cref{tab:deberta_layer_comparison} for \deberta results.} averaged across all tasks.
Our results show that multiple classifiers updating the same weights during fine-tuning diminishes the performance of Early Exit classifiers by \(2.3\%\) on average (\(1.7\%\) for \bert, \(3.0\%
\) for \deberta), with earliest classifiers affected the most (\(3.5\%\) for \bert, \(7.0\%\) for \deberta).
The increased effect on early classifiers supports our hypothesis regarding conflicting gradients: parameters used by early classifiers receive updates from the largest number of classifiers, thus increasing the chance for conflicting gradients and leading to a larger decrease in performance. 

\subsection{Speed-Accuracy Trad-eoff}{\label{speed-accuracy tradeoff}}

So far we observed that \MM classifiers outperform \EE ones. On the other hand, we also note that there is considerable overhead when using a \MM model. For \MM, an instance that makes an exit on classifier \(C_i\), runs all classifiers up to (including) \(C_i\), each being an independent model, while \EE early layer computations are reused by later classifiers. We turn to evaluate the trade-off between these two factors and compare the overall speed-accuracy curve of each adaptive inference method.

We evaluate model scores across different inference speeds using 11 threshold values, evenly spaced in the range \(\left(\frac{1}{\#\text{ of labels}}, 1\right)\).  We note that \(t=\frac{1}{\#\text{ of labels}}\) corresponds to the earliest classifier predicting all labels, while for \(t=1\), the final classifier is used on the entire validation set. 

We compute the speedup ratio for each instance as the number of layers it passes through divided by the number of layers of the full backbone model (12 for BASE models, 24 for LARGE models), and average across all instances. As an  example, for classifiers following layers [1, 4, 6, 12], an instance that exits on the third classifier (i.e., after layer 6), will have a speedup ratio of $\frac{6}{12}$ for \EE and $\frac{11}{12}$ for \MM ($\frac{6}{12} + \frac{4}{12} + \frac{1}{12}$).\footnote{Further experimental details can be found in \cref{sec:appendix_A}.}

\begin{figure}[t]
\centering
\includegraphics[trim={0 0 0 0}, clip, width=\columnwidth]{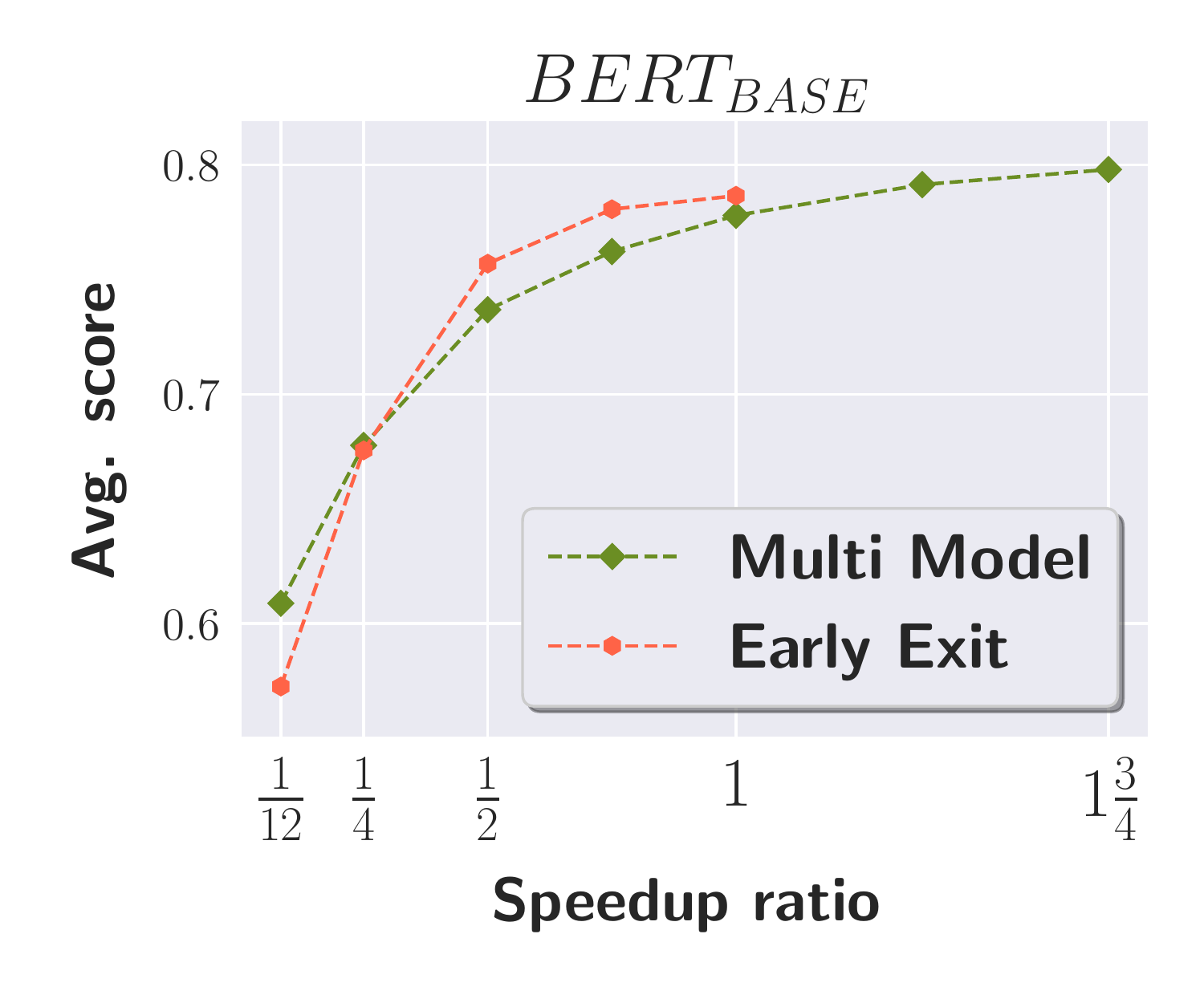}
\caption{\label{EE_MM_speed_acc_comparison}
Speed-accuracy trade-off comparison of \MM  and \EE.
\MM performs better only at fast inference times (up to \(\nicefrac{1}{4}\) of original run time), while \EE dominates the remainder of the range. The graph shows the average task scores (y-axis) as a function of the speedup ratio (x-axis).}
\end{figure}

\begin{figure*}[t]
\centering
\includegraphics[width=\textwidth, trim={0 0 0 0},clip]{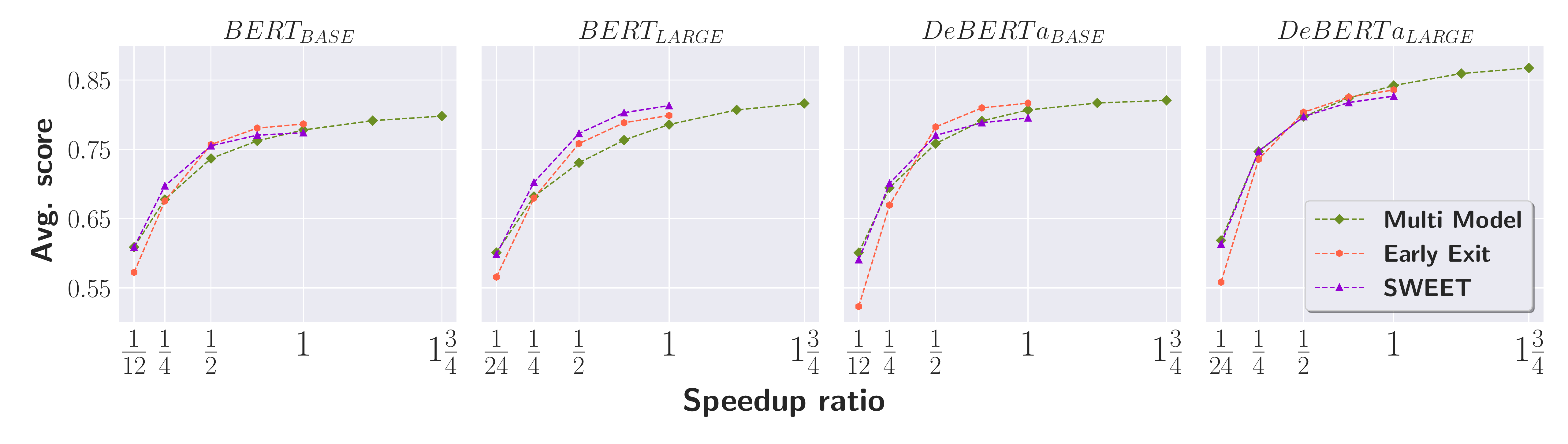}
\caption{Speed-accuracy tradeoff averaged across tasks. \methodname matches the performance of \MM at fast speeds, while maintaining results comparable to \EE at slow speeds.
\label{average_task_results}}
\end{figure*}

We evaluate models on the seven tasks listed in \cref{paragraph: effects CG}, and report the average scores across all tasks. \bertbase results are presented in \cref{EE_MM_speed_acc_comparison}, while the results for all other models can be found in \cref{average_task_results}. 
The speed-accuracy curves reveal two important observations. First, as expected, \MM achieves better scores at fast speeds (when most instances are predicted by early classifiers). Second, although each \MM classifier outperforms its corresponding \EE one, the \MM overhead leads to this approach being outperformed by \EE  at slow speeds (when more instances are predicted by later classifiers).

\section{\methodname: Separating Weights for Early-Exit Transformers}
\label{sec:Method}

Based on our findings, we aim to design an \EE method that takes advantage of the benefits of both \MM and \EE approaches: making the lower \EE classifiers as accurate as the \MM ones, without the additional overhead of the \MM approach.

\subsection{\methodname}

We present \methodname---a method for fine-tuning \EE models that avoids the harmful impact of conflicting gradients. \methodname grants each classifier exclusive control over part of the model's parameters, such that each model parameter is only updated by a single classifier.
This is done by truncating the loss signal from classifier \(C_i\) when reaching the Transformer layer corresponding to classifier \(C_{i-1}\) (\cref{SWEET}, right).

In \methodname, each classification head receives complete control over a portion of the Model's parameters and can alter them in a way that optimizes its own goals.  
Truncating future loss signals makes the very first classifier an independent model, as opposed to a model that shares its parameters with future classifiers. The following classifiers update only a subset of the model's parameters but are affected by early model parameters, allowing them to still make use of earlier computations when processing an instance. We clarify that all model weights are updated simultaneously, as opposed to training the classifiers (and their corresponding layers) sequentially, which would have led to significantly longer fine-tuning, matching that of a \MM model.

 \subsection{A Better Speeds-Accuracy Curve}
We turn to evaluate the speed-accuracy curve of \methodname. 
We use the same experimental setup as in \cref{paragraph: effects CG}; we fine-tune two pre-trained LMs (\bert and \deberta) in two sizes (BASE, LARGE) over the same seven text classification tasks. We use the same exit points and evaluate over the entire validation set using confidence-based exiting. We compare \methodname to two baselines: a standard \EE model and a \MM model. We compute the speedup ratio using the same method as described in \cref{speed-accuracy tradeoff}. For further implementation details, see \cref{sec:appendix_A}.

\begin{figure*}[th]
\centering
\includegraphics[width=\textwidth, trim={0 0 0 0},clip]{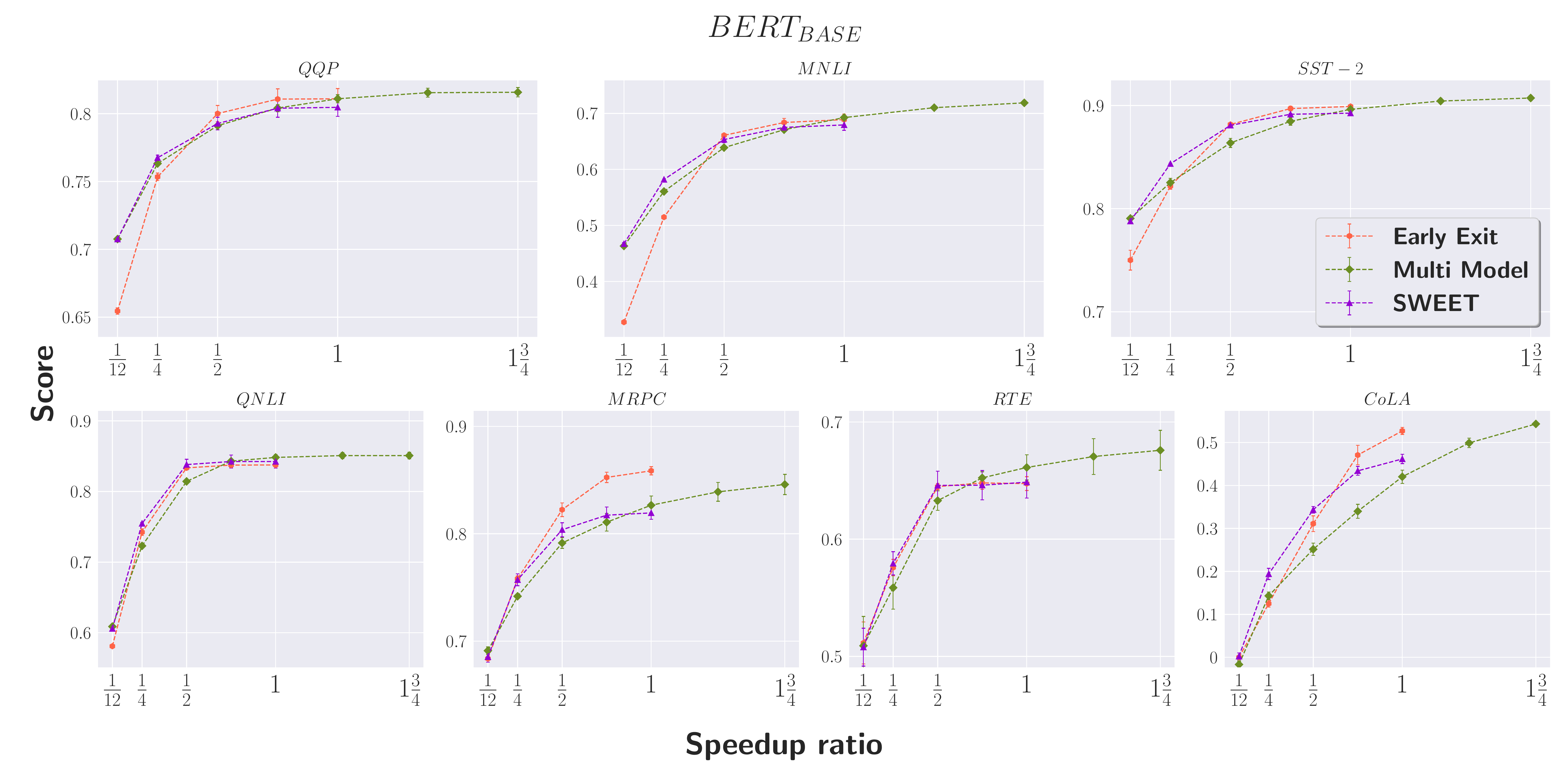}
\caption{Speed-accuracy curve for individual tasks, using a \bertbase model. \methodname outperforms both baselines at fast speeds (lower values of \(x\)) on five out of seven tasks. See \cref{all_model_results} for \bertlarge and \deberta results.\label{bert_base_results}}
\end{figure*}

\cref{average_task_results} presents the speed-accuracy curves of all models, averaged across all tasks. The figure shows that at the fastest speeds, \methodname outperforms \EE for all models and is comparable to, or outperforms, \MM. However, for 3 out of the 4 models (all but \bertlarge), \EE surpasses the performance of \methodname at slow speeds. \methodname's reduced scores at slow inference speeds are likely due to the lower capacity of the later classifiers, stemming from their restricted influence on early model parameters during fine-tuning.

\cref{bert_base_results} presents results on  individual tasks for \bertbase.\footnote{\cref{all_model_results} in \cref{sec:appendix_D} shows individual results for the other models.} On five out of seven tasks, \methodname outperforms both baselines at fast speeds (up to a speedup ratio of \(\nicefrac{1}{2}\)), suggesting that \methodname improves the performance of lower \EE classifiers by avoiding conflicting gradients during training. For two tasks (MRPC and CoLA), \methodname suffers a considerable decrease in performance at slow inference speeds. For the other five tasks \methodname maintains comparable results to \EE.

It is also interesting to examine how \methodname affects \textit{individual} \EE classifiers.
\cref{tab:layer_comparison} shows the results of \bert's individual classifiers trained with \methodname compared to \EE and \MM.\footnote{See \cref{tab:deberta_layer_comparison} in \cref{sec:appendix_C} for \deberta results.} 
\methodname classifiers close much of the gap between \EE and \MM: they outperform those of \EE by \(1.1\%\) on average (\(1.2\%\) for \bert, \(1.0\%\) for \deberta), with the margin being larger for the earliest classifier (\(3.4\%\) for \bert, \(6.2\%\) for \deberta). The final two classifiers trained with \methodname achieve lower scores than those of \EE (\(0.9\% \) on average), probably due to the restricted influence those classifiers have on early model parameters. 
Our results hint that \methodname is able to effectively bridge the gap between \EE and \MM early classifiers, leading to a speed-accuracy curve favoring fast inference speeds.

We note that during our experiments with \EE and \MM models, some models did not converge. In order to ensure the validity of our results, we repeated these experiments with different random seeds, which led to convergence. We emphasize that this did not occur during the training of models using \methodname.

\section{Further Analysis}
\label{sec:Further_Analisys}

 \begin{figure}[t]
\centering
\includegraphics[width=\columnwidth, trim={0 0 0 0},clip]{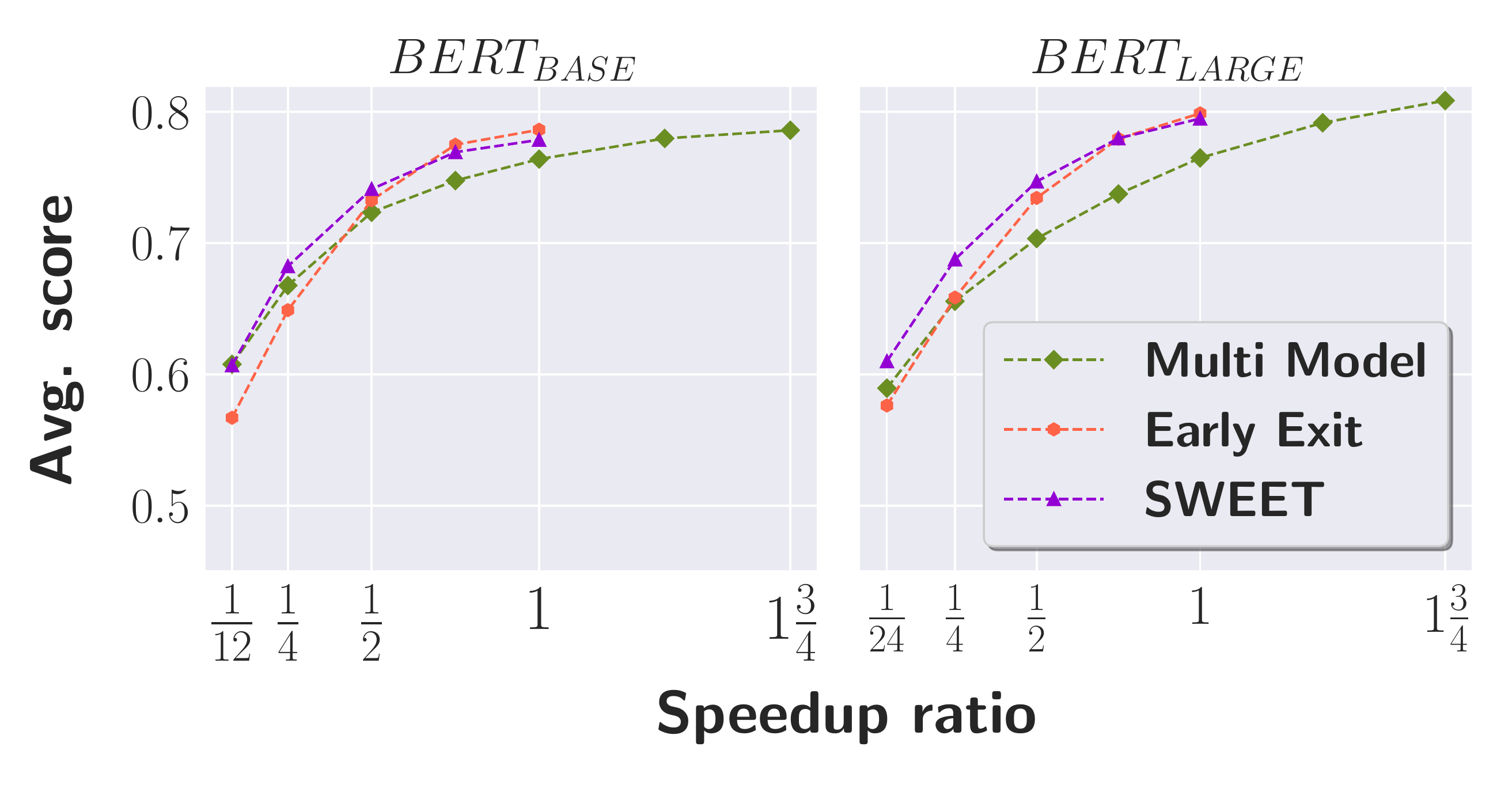}
\caption{\label{average_task_results_lte}Speed-Accuracy curves with the learning-to-exit strategy. As with confidence-based methods, \methodname outperforms both baselines at fast speeds, while maintaining comparable results at slow speeds.}

\end{figure}

\begin{figure*}[t]
\centering
\includegraphics[width=\textwidth, trim={0 0 0 0},clip]{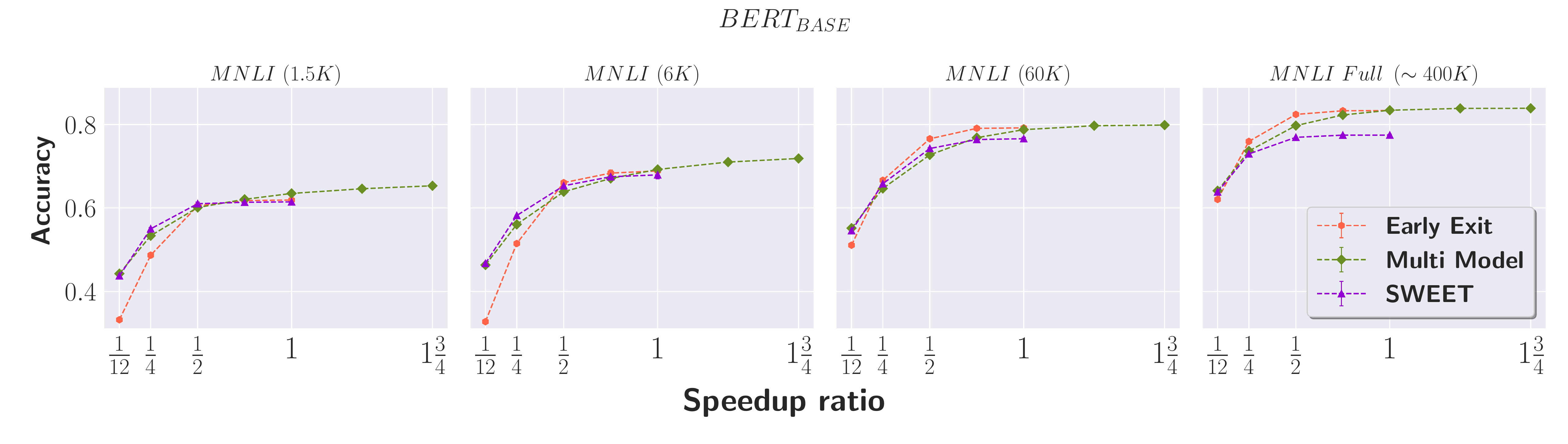}
\caption{\label{varying_data_size}Speed-accuracy curves of 
\bertbase models trained on varying sizes of the MNLI dataset. The title of each plot describes the amount of training data used in the fine-tuning phase. The benefits of using \methodname vanish with large amounts of training data. }

\end{figure*}

We turn to evaluate the robustness of our approach. We start with evaluating \methodname using a different exit strategy. We then test how our results generalize when using varying amounts of training data. 

\paragraph{A different exit strategy: Learning to Exit}\label{Learning to exit}
Our experiments so far have used a confidence-based exit strategy (\cref{sec:Background}). Here we consider a different strategy---learning to exit \cite{xin-etal-2021-berxit}, in which a separate module is trained for each classifier, learning whether the classifier's prediction should be trusted, and an exit should be made. This method also allows for a natural extension to regression problems, unlike confidence-based methods that are limited to classification tasks. 

We use \bert \{BASE, LARGE\} as a backbone model and fine-tune in the same procedure described in \cref{speed-accuracy tradeoff}. Our results (\cref{average_task_results_lte}) reveal that the methods behave similarly to the confidence-based setup;  \methodname outperforms both baselines at the early stage of the curve (fast speeds), while performing on par with \EE at slow speeds. We also measure the performance of individual exit layers, as in \cref{paragraph: effects CG}, for models fine-tuned using learning-to-exit. Our results (\cref{tab:bert_layer_comparison_lte} in  \cref{sec:appendix_E}) reveal a similar behavior to that of models fine-tuned using confidence based early exiting: \MM classifiers surpass the performance of \EE classifiers. Moreover, \methodname's early classifiers outperform those of \EE, while later ones show a slight decrease in performance. 

\paragraph{Varying data sizes} 
Our experiments so far have focused on low-resource setups (a training set of several thousand examples). 
We now evaluate how \methodname performs as we vary the amount of training data. We fine-tune a \bertbase model on different portions of the MNLI dataset using the same experimental setup as in \cref{paragraph: effects CG}. 
Our results, shown in \cref{varying_data_size}, indicate that \methodname is mostly beneficial when training data is limited to a few thousand examples, but still somewhat useful even with as many as 60K training instances. Nonetheless, the effects of conflicting gradients on the earliest \EE classifier tend to disappear when training on the full dataset (400K instances), making it as good as the smallest \MM classifier. Moreover, in that setup, the use of \methodname seems to substantially decrease the performance of later classifiers, suggesting that the harm caused by limiting the influence of classifiers on model parameters may grow with the amount of training data.

\section{Related Work}
\label{sec:Relaterd_work}
Several approaches have been proposed for reducing the inference time of large models~\cite{treviso2022efficient}. 
These include knowledge distillation~\cite{BA_2014distillation,hinton2015distillation}, in which the knowledge of a large teacher model is transferred to a smaller student model; pruning~\cite{lecun1989optimal, frankle2018lottery}, which removes unnecessary model parameters; and weight quantization~\cite{Quantization_Jacob2018, Quantization_Ofir2019}, which reduces the floating point precision of model weights. See (\citealp{Treviso:2023}, Section 6), for a recent survey. Adaptive inference methods studied in this work are an orthogonal approach, which can be used in combination with these methods~\cite{schwartz-etal-2020-right}.

The use of Adaptive inference in deep neural networks has been extensively studied, with successful implementations in various types of networks including recurrent neural networks~\cite{graves2016adaptive} and convolutional neural networks~\cite{teerapittayanon2016branchynet, huang2018multi}. In the context of this work, it has also been applied to existing backbone pre-trained language models:~\citet{xin-etal-2020-deebert} implemented early exit strategies on top of classification tasks.~\citet{zhou2020bert} Introduced a patience-based exiting, requiring sequential classifiers to agree on a label for a prediction to be made.~\citet{xin-etal-2021-berxit} extended the use of \EE in Transformers to regression tasks, as well as addressed the issue of reduced performance of the final classifier through the use of an alternating training algorithm.~\citet{schuster2021consistent} proposed a confidence-based early exit model with guarantees on the agreement between early exit predictions and the final classifier.

~\citet{liu-etal-2022-towards-efficient} presented a strong baseline for efficient NLP by adding multiple classifiers to a \bert model during pre-training. Recently,~\citet{schuster2022confident} adjusted the \EE method to language modeling for text generation, making dynamic computation at the single token level. 

\MM approaches to adaptive inference have been proposed for vision tasks~\cite{enomoro2021learning} as well as for NLP~\cite{li2020cascadebert, Varshney:2022}.~\citet{mamou2022tangobert} introduced a two-tier design with a highly efficient model serving as the first model and a powerful model serving as the second, enabling the possibility of achieving extremely fast inference speed. 

Finally, the concept of conflicting gradients has been mainly studied in the context of multi-task learning, where a single model is faced with solving different tasks~\cite{yu2020gradient, liu2021conflict}. To the best of our knowledge, no previous work examined this in the context of \EE.

\section{Conclusion}
\label{sec:Conclusion}
In this work, we analyzed the performance of two common adaptive inference methods--\EE and \MM. We found evidence that model weights are updated by conflicting gradients in the training process of \EE models, causing  classifiers to perform at a sub-optimal level. Despite this, we showed that regarding the entire speed-accuracy curve, \EE is still favorable to \MM due to the overhead of using independent model runs in a \MM setup. 

To address these findings, we proposed \methodname, a novel \EE method, which avoids conflicting gradients by allocating each \EE classifier a subset of model weights which are updated solely by it. We found that for \EE models trained with  \methodname, early classifiers perform better than those of standard \EE, but later classifiers of \methodname are not as good. These measures lead to \methodname outperforming both \EE and \MM in fast speeds, with slightly worse results than \EE at slow speeds. 

Overall, our results demonstrate that \EE models can benefit from fine-tuning algorithms that are tailored to their architecture, and that SWEET is a promising approach for improving the speed-accuracy trade-off of \EE models in the context of adaptive inference.

\section{Limitations}
\label{sec:Limitations}
This work focuses on the effects of adaptive inference in a low-resource setting, specifically when training data is limited. Our experiments (\cref{sec:Further_Analisys}) suggest that the negative impact of conflicting gradients may be less prominent when larger amount of training data is available. 

Our experiments were conducted using relatively small pre-trained language models (\(\leq 350M\) parameters) due to computational constraints, and we defer the replication of our findings with larger, more powerful models to future work. Nonetheless, our results have important implications for the growing trend of increasingly large language models. We hope this work inspires further research on methods to reduce the computational cost of NLP.

This work concentrates on evaluating the speed-accuracy trade-off of \MM and \EE at inference time. We recognize that there are additional factors, such as memory usage, batch processing, and training duration, that could be considered when comparing these methods.

Finally, we experimented with seven text classification tasks in English. We recognize that results may vary for other tasks and languages.

\section*{Acknowledgements}
\label{sec:Acknowledgements}
We acknowledge Yarden Tal for her early contributions to this work, and Gabriel Stanovsky for his advice and meaningful feedback.  This work was supported in part by the Israel Science Foundation (grant no. 2045/21), NSF-BSF grant 2020793, and by a grant from Intel Labs.

\bibliography{custom}
\bibliographystyle{acl_natbib}

\clearpage

\appendix

\section{Implementation Details}
\label{sec:appendix_A}

\begin{table*}[t] 
\centering
\setlength{\tabcolsep}{9.4pt}
\begin{tabularx}{\textwidth}{ lcccccccc}
\toprule
 \textbf{Model} & \textbf{Size} &  \multicolumn{7}{c}{\textbf{Task}}\\
 \midrule\rule{0pt}{2.0ex}
&&\textbf{SST-2} & \textbf{MRPC} & \textbf{CoLA} & \textbf{MNLI} & \textbf{QQP} & \textbf{QNLI} & \textbf{RTE}\\
\cmidrule{3-9}
\multirow{2}{*}{\bert} 
&BASE &
\(\text{5}\backslash\text{5}\backslash\text{5}\)&
\(\text{5}\backslash\text{3}\backslash\text{4}\) &
\(\text{4}\backslash\text{4}\backslash\text{5}\) &
\(\text{5}\backslash\text{5}\backslash\text{5}\) &
\(\text{5}\backslash\text{5}\backslash\text{5}\) &
\(\text{5}\backslash\text{4}\backslash\text{5}\) &
\(\text{5}\backslash\text{5}\backslash\text{4}\) \\
&LARGE & 
\(\text{4}\backslash\text{2}\backslash\text{2}\)&
\(\text{4}\backslash\text{5}\backslash\text{4}\)&
\(\text{4}\backslash\text{3}\backslash\text{4}\)&
\(\text{4}\backslash\text{3}\backslash\text{4}\)&
\(\text{5}\backslash\text{3}\backslash\text{4}\)&
\(\text{4}\backslash\text{4}\backslash\text{3}\)&
\(\text{3}\backslash\text{3}\backslash\text{3}\)\\
\midrule
\multirow{2}{*}{\deberta}
&BASE & 
\(\text{3}\backslash\text{3}\backslash\text{3}\)&
\(\text{4}\backslash\text{5}\backslash\text{5}\)&
\(\text{4}\backslash\text{5}\backslash\text{4}\)&
\(\text{2}\backslash\text{4}\backslash\text{3}\)&
\(\text{4}\backslash\text{5}\backslash\text{5}\)&
\(\text{3}\backslash\text{4}\backslash\text{3}\)&
\(\text{2}\backslash\text{1}\backslash\text{4}\)\\

&LARGE & 
\(\text{2}\backslash\text{4}\backslash\text{3}\)&
\(\text{3}\backslash\text{2}\backslash\text{1}\)&
\(\text{3}\backslash\text{3}\backslash\text{2}\)&
\(\text{2}\backslash\text{2}\backslash\text{2}\)&
\(\text{2}\backslash\text{2}\backslash\text{3}\)&
\(\text{3}\backslash\text{2}\backslash\text{1}\)&
\(\text{1}\backslash\text{2}\backslash\text{2}\)\\
\bottomrule
\end{tabularx} 
\caption{\label{tab:LR_selection} 
Chosen initial learning rate to optimize area under 
 the speed-accuracy curve of each model, size, task. All numbers should be multiplied by 1e-5.  \(x\backslash y\backslash z\) represent the initial learning rate of \EE\textbf{\textbackslash} \MM\textbf{\textbackslash} \methodname  respectively.}
\end{table*}

\paragraph{Further implementation details} 
 For fine-tuning \bert models, we use a batch size of 16. For fine-tuning \deberta, due to GPU memory constraints, we use a batch size of 16 for BASE and 8 for LARGE. We fine tune the models for two epochs with a maximal sequence length of 256. We use \(\beta= 0.9, 0.999\) for the \textit{AdamW} optimizer with linear LR-decay. We optimize the initial learning rate for each  method, model size \& task by performing a search over five values \{1e-5, 2e-5, 3e-5, 4e-5, 5e-5\} and choose the value leading to the largest area under the speed-accuracy curve. Chosen LRs are presented in table \cref{tab:LR_selection}

 All experiments were run on a single NVIDIA A5000 GPU. The overall computational budget was \({\sim}1000\) GPU hours.  We implement all methods using  the HuggingFace Transformer library \cite{wolf-etal-2020-transformers}.

\paragraph{Speedup evaluation}
The speed-up ratio of the model when using exit threshold \(t\) is calculated using the formula: 
\begin{equation}
\text{speedup}_t = \frac{\sum_{i=1}^M S_i^t\cdot L_i}{ L_M\cdot\sum_{i=1}^M S_i^t}
\end{equation}
where \(M\) is the number of classifiers used for the model, \(L_i\) denotes the number of the layer preceding the \(i\)-th classification head and \(S_i^t\) denotes the number of samples classified by the \(i\)-th classifier when using threshold \(t\).
 
The same exit threshold can lead to different speedup ratios amongst  models trained with different random seeds.  We use linear interpolation to approximate the accuracy score at set speedup ratios. We evaluate at \(\left(\nicefrac{1}{12}, \nicefrac{1}{4}, \nicefrac{1}{2},\nicefrac{3}{4}, 1\right)\)\footnote{For LARGE models we evaluate at \(\nicefrac{1}{24}\) as the fastest speedup ratio.} and report the average across three random seeds as well as a \(95\%\) confidence interval.\footnote{For \MM we also use \(\left(1\nicefrac{3}{8}, 1\nicefrac{3}{4}\right) \) as overhead causes the model to perform at such "speedup" ratios.} We use temperature scaling \cite{guo2017calibration} to make classifiers confidence, and therefore early-exiting decisions, more reliable. Note that this scaling is monotonic and therefore does not influence predictions.

\section{Conflicting Gradients}
\label{sec:appendix_B}
\begin{figure}[h]
\centering
\includegraphics[trim={0 0 0 0}, clip, width=\columnwidth]{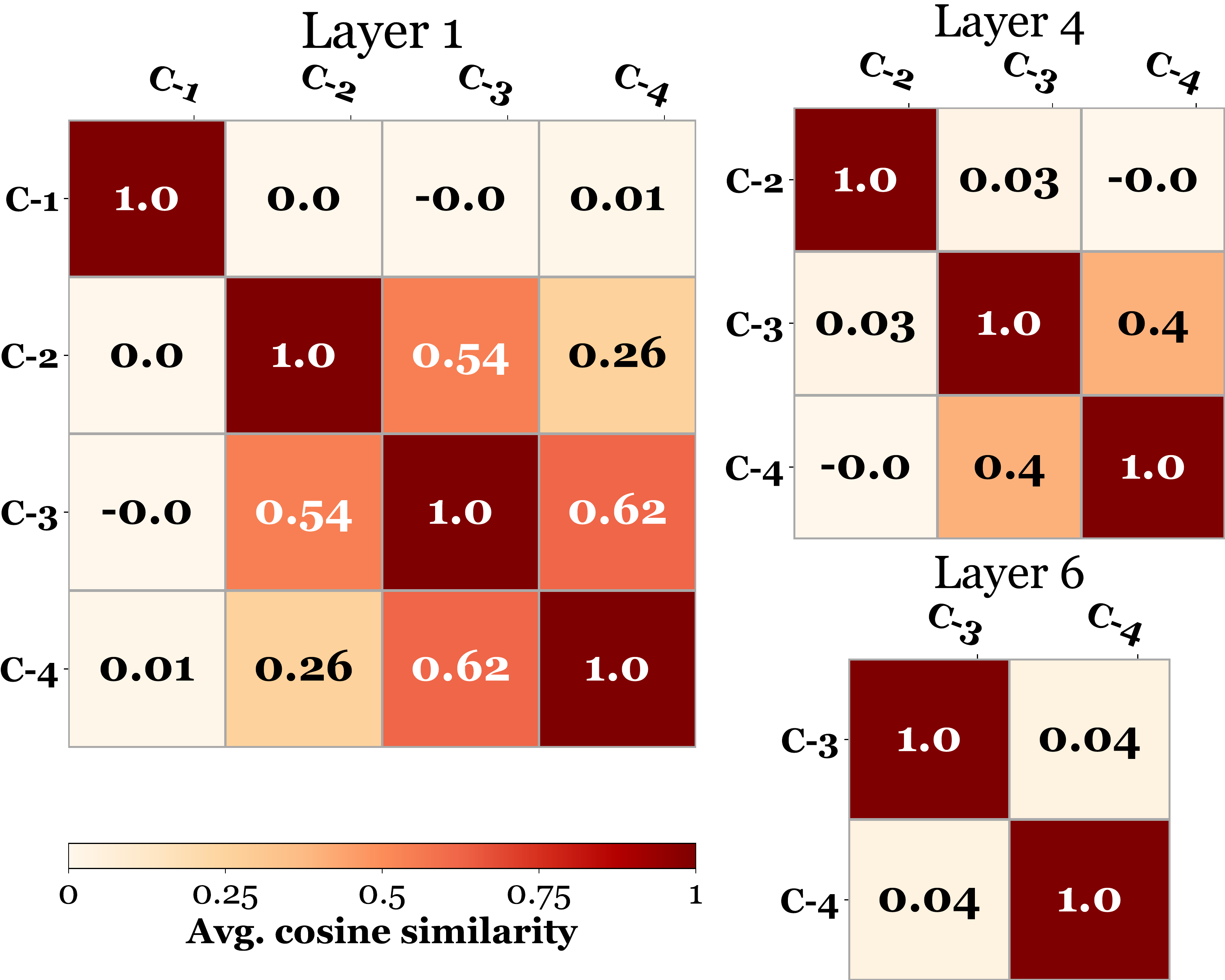}
\caption{\label{gradient_similarity_second_batch}
Average cosine similarity between classifiers' gradient updates of model layers. C-i stands for Classifier \(i\). Layer 1 (preceding C-1) is updated by 3 classifiers, while layers 4 (preceding  C-2), and 6 (C-3) are updated by 3 \& 2 classifiers respectively.
For each layer, the gradient update of the following classifier is roughly orthogonal to those of future classifiers, whereas gradient updates of higher classifiers tend to better align with one another.}
\end{figure}

We replicate the experiment done in \cref{conflicting_gradients} with another batch of size 16 to examine if our findings  generalize.  results  presented in \cref{gradient_similarity_second_batch} show a similar trend to \cref{gradient_similarity}: Gradient updates of current classifiers are roughly orthogonal to those of future classifiers, whereas future classifier updates are more aligned. 

\section{\deberta Individual Layer Comparison}
\label{sec:appendix_C}

\cref{tab:deberta_layer_comparison} Shows individual classifier results for \deberta models. As with the \bert results, \MM classifiers outperform corresponding \EE classifiers. \methodname early classifiers are better than \EE ones, while later classifiers tend to downgrade in performance.

\setlength{\tabcolsep}{2.3pt} 

\begin{table}[ht] 
\centering

\begin{tabularx}{\columnwidth}{@{} lccccc@{}}
\toprule
 \textbf{Size}&\textbf{Method} & \multicolumn{4}{c}{\textbf{Exit Layer}}\\[0.3ex]
 \midrule\rule{0pt}{2.0ex}
&&\textbf{1} & \textbf{4} & \textbf{6} & \textbf{12}\\
\cmidrule{3-6}
\multirow{3}{*}{BASE} & MM &  \(\textbf{60.1}_{0.4}\) & \(\textbf{74.6}_{0.6}\) & \(\textbf{78.3}_{1.1}\) & \(\textbf{82.2}_{0.7}\) \\
& EE & \(52.3_{0.4}\) & \(70.9_{0.6}\) & \(77.9_{0.8}\) & \(81.7_{0.1}\) \\
& \methodname & \(59.1_{1.5}\) & \(72.4_{0.2}\) & \(75.1_{1.1}\) & \(79.5_{0.5}\) \\[1ex]
\midrule
&&\textbf{1} & \textbf{6} & \textbf{12} & \textbf{24}\\
\cmidrule{3-6}
\multirow{3}{*}{LARGE}
& MM &  \(\textbf{61.9}_{0.1}\) & \(\textbf{76.2}_{0.4}\) & \(\textbf{80.2}_{0.9}\) & \(\textbf{86.8}_{0.2}\)\\
&EE & \(55.8_{0.9}\) & \(74.7_{1.1}\) & \(79.2_{0.8}\) & \(83.6_{1.9}\) \\
& \methodname & \(61.3_{0.7}\) & \(76.0_{0.5}\) & \(77.8_{0.4}\) & \(82.7_{0.4}\) \\
\bottomrule\bottomrule

\end{tabularx} 
\caption{\label{tab:deberta_layer_comparison} Results of individual classification layers averaged over all tasks using \deberta as a backbone model. Best scores are highlighted in bold, standard deviation (across random seeds) is reported in subscript. The results for BERT models presented in \cref{tab:layer_comparison}.}
\end{table}

\section{Individual Task Results}
\label{sec:appendix_D}

\begin{figure*}[ht]
\centering
\begin{subfigure}[b]{\textwidth}
\includegraphics[width=\textwidth, trim={0 0 0 0},clip]{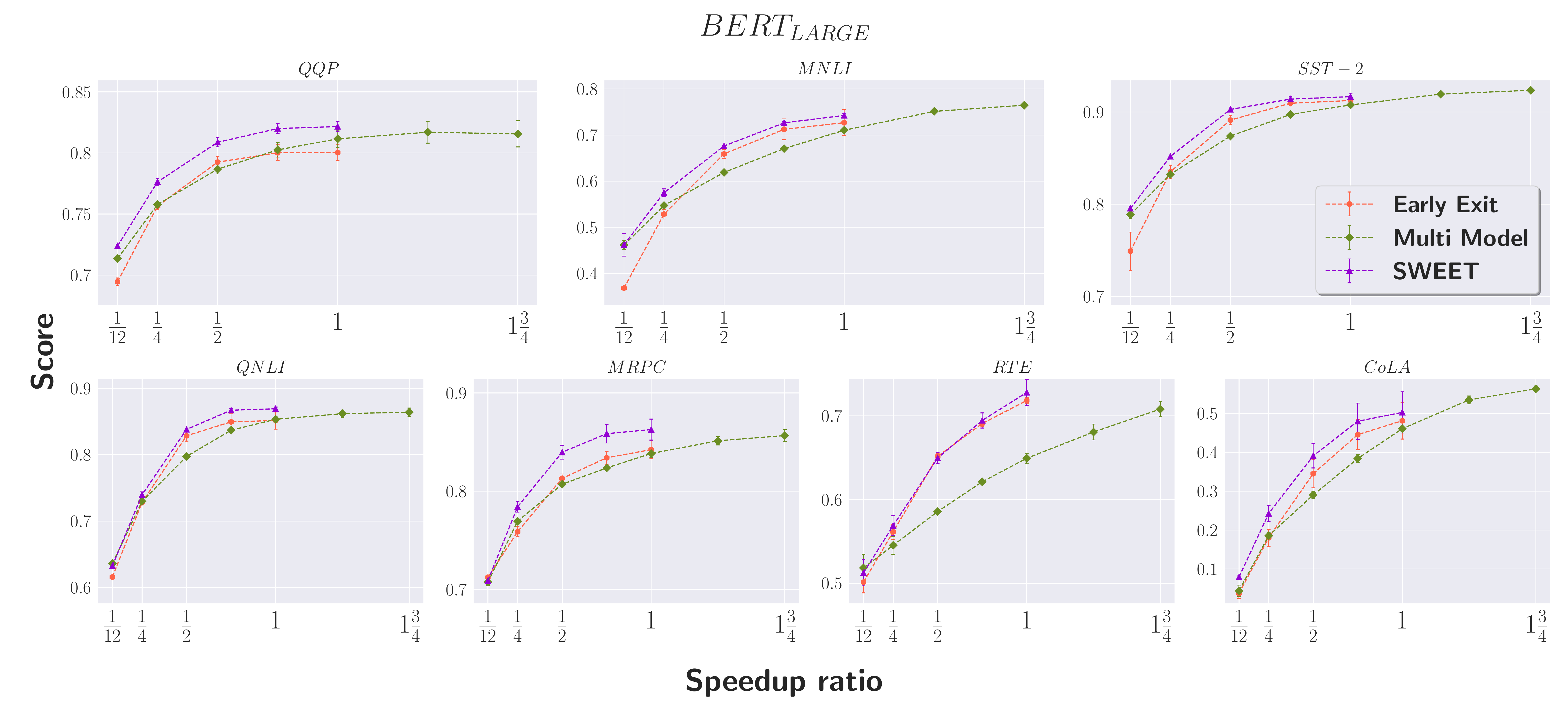}
\end{subfigure}

\begin{subfigure}[b]{\textwidth}
\centering
\includegraphics[width=\textwidth, trim={0 0 0 0},clip]{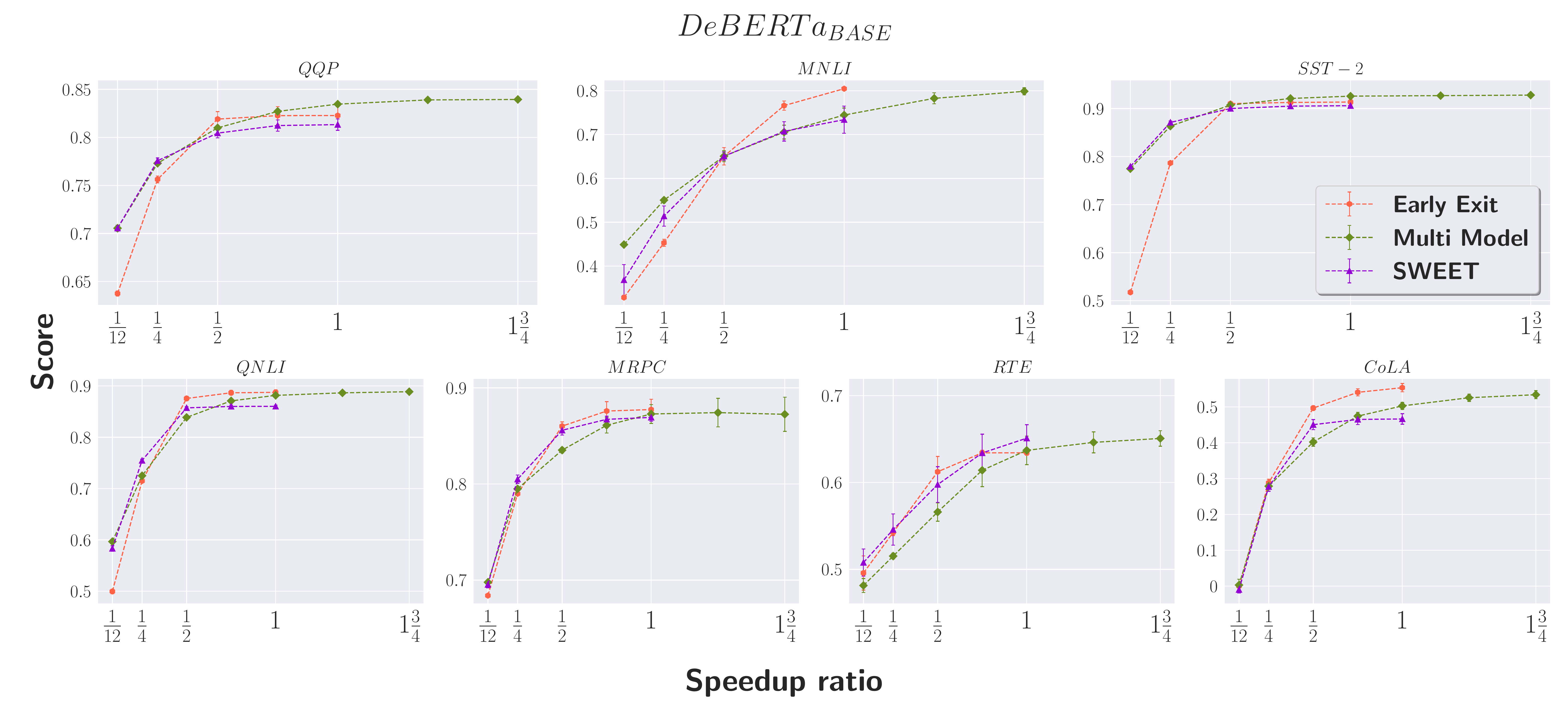}

\end{subfigure}

\begin{subfigure}[b]{\textwidth}
\centering
\includegraphics[width=\textwidth, trim={0 0 0 0},clip]{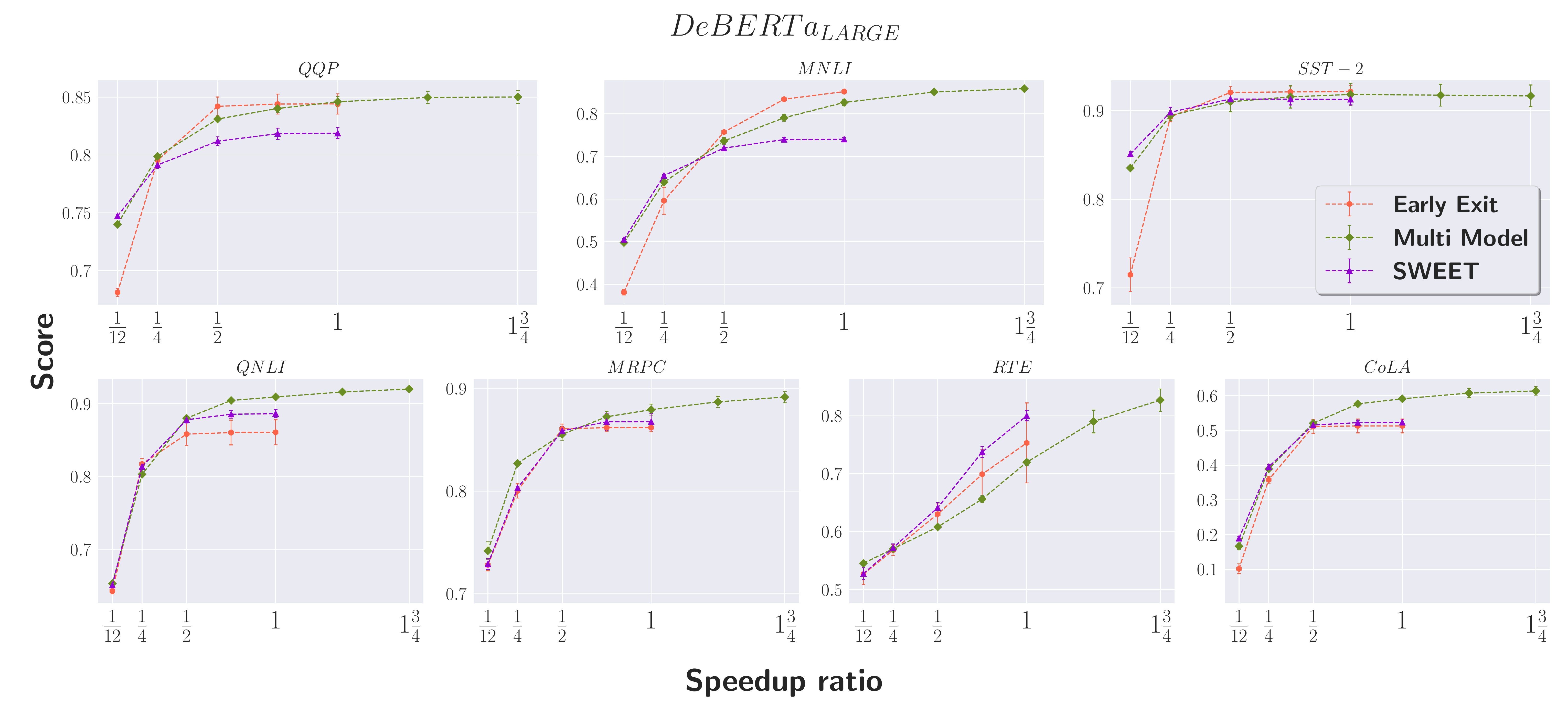}
\end{subfigure}
\caption{Speed-accuracy curves for individual tasks. For \bertlarge, \methodname outperforms both baselines across the entire curve. For \deberta, \methodname performs better than \EE at high speeds on 9 out of 14 experiments. \label{all_model_results}}

\end{figure*}

\cref{all_model_results} shows results on individual tasks for \bertlarge and \deberta (BASE \& LARGE). For \bertlarge, \methodname outperforms both baselines throughout the entire speed-accuracy curve over all tasks examined. For \deberta models, results are similar to those of \bertbase, where \methodname performs better at high speeds (small speedup ratio) and is dominated at low speeds (where later classifiers do most of the heavy lifting).  

\section{Learning-to-exit Individual Layer Comparison}
\label{sec:appendix_E}
\cref{tab:bert_layer_comparison_lte} Shows individual classifier results for \bert models fine tuned with the learning to exit strategy. As with the confidence-based results, \MM classifiers outperform corresponding \EE classifiers. \methodname early classifiers are better than \EE ones, while later classifiers tend to downgrade in performance.

\setlength{\tabcolsep}{1.75pt} 

\begin{table}[H] 
\centering

\begin{tabularx}{\columnwidth}{@{} lccccc@{}}
\toprule
 \textbf{Size}&\textbf{Method} & \multicolumn{4}{c}{\textbf{Exit Layer}}\\[0.3ex]
 \midrule\rule{0pt}{2.0ex}
&&\textbf{1} & \textbf{4} & \textbf{6} & \textbf{12}\\
\cmidrule{3-6}
\multirow{3}{*}{BASE} & MM &  \(\textbf{53.7}_{0.5}\) & \(\textbf{66.1}_{0.4}\) & \(\textbf{71.1}_{0.3}\) & \(\textbf{75.4}_{2.7}\) \\
& EE & \(49.5_{1.1}\) & \(63.0_{0.2}\) & \(70.0_{0.6}\) & \(75.3_{0.4}\) \\
& \methodname & \(53.5_{0.4}\) & \(64.9_{0.9}\) & \(68.5_{1.4}\) & \(74.1_{0.4}\) \\[1ex]
\midrule
&&\textbf{1} & \textbf{6} & \textbf{12} & \textbf{24}\\
\cmidrule{3-6}
\multirow{3}{*}{LARGE}
& MM &  \(51.6_{0.3}\) & \(59.9_{0.4}\) & \(67.6_{2.6}\) & \(\textbf{77.9}_{0.8}\)\\
&EE & \(50.5_{0.8}\) & \(59.3_{0.9}\) & \(68.8_{1.3}\) & \(76.6_{1.6}\) \\
& \methodname & \(\textbf{53.9}_{0.5}\) & \(\textbf{60.6}_{0.2}\) & \(\textbf{69.5}_{0.4}\) & \(75.81_{1.0}\) \\
\bottomrule\bottomrule

\end{tabularx} 
\caption{\label{tab:bert_layer_comparison_lte} Results of individual classification layers averaged over all tasks using \bert as a backbone model, fine-tuned using learning-to-exit.  Best scores are highlighted in bold, standard deviation (across random seeds) is reported in subscript.}
\end{table}


\end{document}